\documentclass[10pt,twocolumn,letterpaper]{article}

\usepackage[accsupp]{axessibility}  
\usepackage[pagenumbers]{cvpr} 

\definecolor{cvprblue}{rgb}{0.21,0.49,0.74}
\usepackage[pagebackref,breaklinks,colorlinks,allcolors=cvprblue]{hyperref}
\usepackage{makecell}
\def\paperID{7} 
\def\confName{CVPR}
\def\confYear{2025}

\title{A Visual RAG Pipeline for Few-Shot Fine-Grained Product Classification}

\author{Bianca Lamm\\
Markant Services International GmbH\\
{\tt\small Bianca.Lamm@de.markant.com}
\and
Janis Keuper\\
IMLA, Offenburg University and \\University of Mannheim\\
{\tt\small keuper@imla.ai}
}

\begin{document}
\maketitle
\setlength{\fboxsep}{0pt} 
\begin{abstract}
\noindent Despite the rapid evolution of learning and computer vision algorithms, Fine-Grained  Classification (FGC) still poses an open problem in many practically relevant applications. 
In the retail domain, for example, the identification of fast changing and visually highly similar products and their properties are key to automated price-monitoring and product recommendation.\\
This paper presents a novel Visual RAG pipeline that combines the Retrieval Augmented Generation (RAG) approach and Vision Language Models (VLMs) for few-shot FGC.
This Visual RAG pipeline extracts product and promotion data in advertisement leaflets from various retailers and simultaneously predicts fine-grained product ids along with price and discount information.
Compared to previous approaches, the key characteristic of the Visual RAG pipeline is that it allows the prediction of novel products without re-training, simply by adding a few class samples to the RAG database.\\
Comparing several VLM back-ends like \textit{GPT-4o} \cite{OpenAI2024GPT4o}, \textit{GPT-4o-mini} \cite{OpenAI_GPT4omini}, and \textit{Gemini 2.0 Flash} \cite{googleGeminiModels}, our approach achieves \textit{86.8\%} accuracy on a diverse dataset.

\end{abstract}
\begin{figure*}
    \centering
    \includegraphics[width=\linewidth]{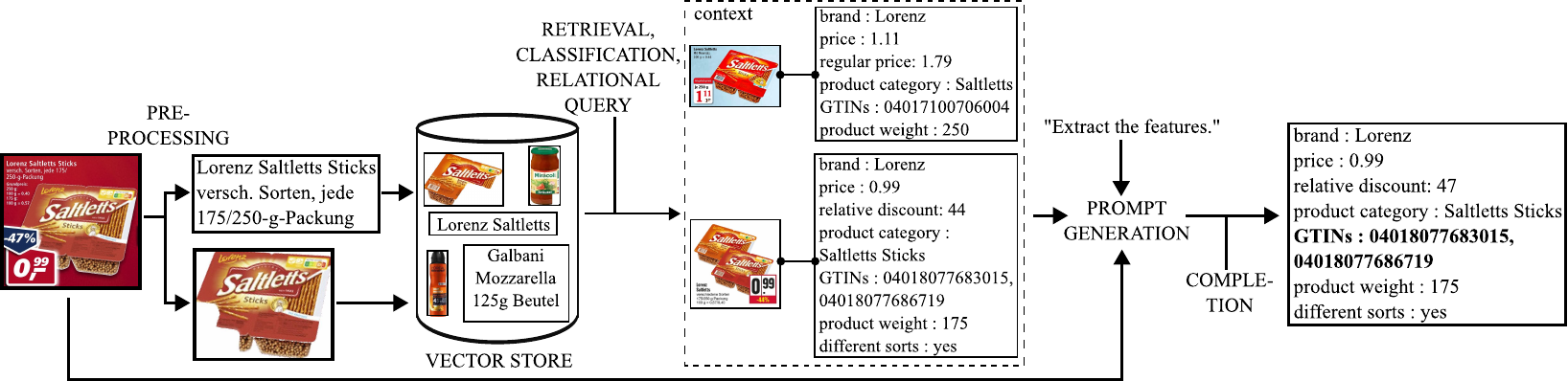}
    \caption{Illustration of the presented Visual RAG pipeline.
    The pipeline is based on the RAG approach and is characterized by five main steps: \textit{Preprocessing}; \textit{Vector Store}; \textit{Retrieval, Classification, Relational Query}; \textit{Prompt Generation}; and \textit{Completion}.
    Moreover, a contextual knowledge comprising few-shot samples with corresponding task solutions is appended to the prompt for the employed VLM.
    The prediction of the target \textit{GTINs} serves as FGC.
    The additional predictions deliver to enrich the objectives.
    }
    \label{fig:visual_abstract}
\end{figure*}    
\section{Introduction}
\label{sec:intro}
\noindent The task of Fine-Grained Classification (FGC) enables a detailed analysis and more precise categorization of items that are highly similar in nature.
The differentiation of items arises in several areas, such as animals, plants, cars, and retail products \cite{wei2021fine}.
In the retail domain, traded products are continuously changing, while thousands of products are sold in a single supermarket.
Identical products are identified by the standardized and unique Global Trade Item Number (GTIN) \cite{gs1GlobalTrade}. Hence, the prediction of GTINs is defined as a fine-grained classification task.
Products differ in a fine-grained manner, \eg, due to similar content but different packaging size.
The previous methods used for FGC are mostly based on Convolutional Neural Networks or Vision Transformers \cite{dosovitskiy2021imageworth16x16words}, which provide a reasonable accuracy but have to be retrained very often in practice in order to keep up with the fast changing product portfolios.
In recent years, Vision Language Models (VLMs) have significantly advanced the integration of visual and textual data, enabling more sophisticated multi-modal understanding \cite{zhang2024visionlanguagemodelsvisiontasks}.
These models have the capability for solving complex reasoning tasks, including image captioning and visual question answering \cite{zhang2024visionlanguagemodelsvisiontasks, wang2024largescalemultimodalpretrainedmodels}.
So far, VLMs have mainly a knowledge about public available data \cite{zhang2024visionlanguagemodelsvisiontasks}, which typically does not include detailed information about retail products.
The Retrieval Augmented Generation (RAG) approach shows promising outcomes by incorporating external knowledge sources \cite{gao2024retrievalaugmentedgenerationlargelanguage}.
Recently, the emergence of multi-modal inputs for RAG is an ongoing research topic  \cite{abootorabi2025askmodalitycomprehensivesurvey}.
The creation of a custom database which is required for the RAG method offers the ability to provide the necessary context to VLMs for specific tasks.\\
\\
In this paper, we investigate the combination of multi-modal RAG with VMLs for a few-shot fine-grained classification of retail products and their properties from advertisement leaflets. 
Our approach allows to add novel products to the FGC pipeline, simply by adding a few samples to the RAG database. 
This re-training free approach achieves state-of-the-art accuracies on a multi-modal benchmark.
Our introduced Visual RAG pipeline is a novel approach to RAG using VLMs.
The creation of a context, which is part of the VLM prompt, is essential for facilitating the VLMs comprehension of the task.
In addition, the investigation of the RAG approach on visual data from the retail domain constitutes a novel methodology toward FGC.
\section{Related Work}
\label{sec:related_work}
\noindent An overview of the topics FGC, image datasets in retail, and RAG is presented in the \Cref{subsec:related_work_fine_grained_classification,subsec:related_work_image_dataset_retail,subsec:related_work_retrieval_augmented_generation}.

\subsection{Fine-Grained Classification}
\label{subsec:related_work_fine_grained_classification}

The task of FGC is presented in different domains in particular on image classification.
An example is the image classification of bird species \cite{wei2021fine}.
\cite{liu2024yub} provides a dataset of 120k images split in 60 classes.
Further information about bounding boxes, segmentation masks, or habitat environment are provided.
The authors of \cite{krause2013collecting} investigate the classification of car models.
This image dataset contains about 16k images divided in 197 classes.
The image dataset LZUPSD \cite{yuan2024dataset} consists of about 4.5k images.
The images show plant seeds from 88 different seeds.

\subsection{Image Datasets in Retail}
\label{subsec:related_work_image_dataset_retail}

There are thousands of retail products among which individuals are difficult to distinguish due to, \eg, the packaging.
The work of \cite{brosch2023data} presents a dataset consisting of about 1k images.
The images were taken under studio conditions and show single food products from different views.
In addition, class labels and object detection labels are provided.
The Grocery Store Image Dataset \cite{klasson2019hierarchical} combines retail products and the task of FGC.
The objects on the images of the dataset are fruits and vegetables as well as dairy and juice products. 
The dataset contains about 5.1k images divided in 81 fine-grained classes.
Per class an iconic image and multiple natural images are collected.
The authors of \cite{zheng2021semantic} provide an image dataset that comprises the pages of IKEA catalogs.
On basis of this data, the IKEA dataset \cite{alfassy2022fetaspecializingfoundationmodels} is created.
The dataset contains about 9.5k images and almost 24k texts extracted from the aforementioned pages.
The Retail-786k \cite{lamm2024retail786klargescaledatasetvisual} image dataset consists of about 786k images split into about 3k classes.
The images show product advertisements cropped from leaflet pages.
This dataset forms the basis of the investigations of this paper.

\subsection{Retrieval Augmented Generation}
\label{subsec:related_work_retrieval_augmented_generation}

The RAG method derives from the Natural Language Processing (NLP) \cite{lewis2020retrieval}.
This approach combines data retrieval and text generation by accessing and using external knowledge \cite{lewis2020retrieval}.
Recently, multi-modal RAG approaches extend the classic RAG method by using other modalities, such as text, images, audio, or video.
A comprehensive overview of datasets and benchmarks that evaluate multi-modal RAG approach is shown in \cite{abootorabi2025askmodalitycomprehensivesurvey}.
\cite{yasunaga2023retrievalaugmentedmultimodallanguagemodeling} introduces the retrieval and generation of text and images.
\cite{yu2024visragvisionbasedretrievalaugmentedgeneration} presents a VLM-based RAG pipeline, called VisRAG.
Furthermore, the authors of \cite{rao2024ravenmultitaskretrievalaugmented} expand the query input to the VLM with the retrieved text and image samples.
The improvement of the retrieved samples is evaluated in \cite{chen2024mllmstrongrerankeradvancing} by suggesting a knowledge-enhanced reranking and noise-injected training.
\section{Dataset}
\label{sec:dataset}
\noindent We use a subset of the \textit{Retail-786k} \cite{lamm2024retail786klargescaledatasetvisual} image dataset, supplemented with additional textual data per image.
The addition of textual data to the image dataset enables more complex investigations to be carried out.
\Cref{subsec:image_data} provides a detailed description of the images.
In \Cref{subsec:product_promotion_data}, the thorough explanation of the text information are presented.
The unique characterization of dataset used is the incorporation of textual data pertaining to products and promotions.
\Cref{fig:example_data_of_dataset} illustrates an item of the dataset that consists of an image and its product and promotion data.
The dataset is published on the website: \url{https://huggingface.co/datasets/blamm/retail_visual_rag_pipeline}.

\subsection{Image Data}
\label{subsec:image_data}
The dataset used consists of 4,771 images labeled into 367 different classes.
The images are split into subsets of 3,670 training images and 1,101 test images.
The dataset is balanced but the high number of classes with relatively few examples per class makes the FGC task very challenging.
Each training class has 10 images, each test class has 3 images. 
The images have a size of 512 pixels on the longer edge (see \Cref{subfig:example_image_dataset} for an example).

\subsection{Product and Promotion Data}
\label{subsec:product_promotion_data}
\begin{figure}
    \begin{minipage}{\linewidth}
        \centering
        \fbox{\includegraphics[width=0.5\linewidth]{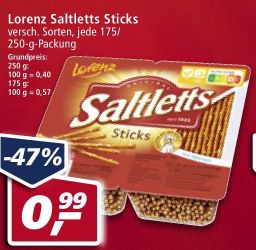}}
        \subcaption{Image from the dataset.}
        \label{subfig:example_image_dataset}
    \end{minipage}
    \par\vspace{18pt}
    \begin{minipage}{\linewidth}
        \centering
        \small 
        \begin{tabular}{lll}
            \hline
            data type   &target    &value\\
            \hline
            product
            &brand              &Lorenz\\
            &product category   &Saltletts Sticks\\
            &GTINs              &04018077683015,\\
            &                   &04018077686719\\
            &product weight     &175.0 Gramm\\
            &different sorts    &yes\\
            promotion
            &price                  &0.99\\
            &regular price          &NaN\\
            &relative discount      &47\\
            &absolute discount      &NaN\\
            \hline
        \end{tabular}
        \subcaption{Product and promotion data.
        Missing target values are stored as NaN.}
        \label{tab:product_promotion_data}
    \end{minipage}
    \caption{Illustration of an item in the dataset that consists of an image (\ref{subfig:example_image_dataset}) and textual product and promotion data (\ref{tab:product_promotion_data}).}
\label{fig:example_data_of_dataset}
\end{figure}

For additional details, the data about the product and promotion is made accessible. The product information comprises a detailed account of product properties: \textit{brand}, \textit{product category}, \textit{GTINs}, \textit{product weight}, and \textit{different sorts}. In the case that a promotion covers a variety of different types or flavors of the product, the GTIN of each type is recorded.
It is common practice for promotional pricing to include not only the special-offer price of the product, but also the regular price and/or the discount. Discounts can be classified as either relative or absolute. Hence, the characterization of a promotion includes the prediction targets: \textit{price}, \textit{regular price}, and \textit{relative discount} or \textit{absolute discount} additional to the FGC of the product GTINs (see \Cref{tab:product_promotion_data} for an example of product and promotion data).
The importance of the target \textit{GTINs} is given by the unique identifier for products.
A promotion image can have assigned either a single GTIN or multiple GTINs.
A multitude of GTINs originates by the promotion of products with different flavors or different weight quantity.
In the retail and supplier domain, the data about GTINs are essential for reporting and analysis.

\subsection{Fine-Grained Dataset}
\label{subsec:dataset_fine_grained}

\begin{figure}
    \centering
    \begin{subfigure}[c]{0.48\linewidth}
        \centering
        \fbox{\includegraphics[width=\linewidth]{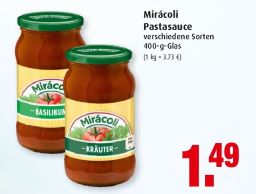}}
        \caption{Product weight is 400g.}
        \label{subfig:fine_grained_image_weight_400}
    \end{subfigure}\hfill
    \begin{subfigure}[c]{0.48\linewidth}
        \centering
        \fbox{\includegraphics[width=\linewidth]{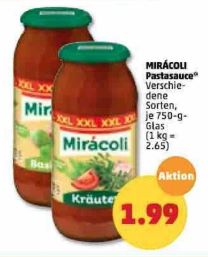}}
        \caption{Product weight is 750g.}
        \label{subfig:fine_grained_image_weight_500}
    \end{subfigure}
    \caption{Illustrations of images that demonstrate a fine-grained difference due to variations in product weight.
    The evaluations of ResNet50 \cite{he2016deep} and BERT \cite{kenton2019bert} show misclassification of these images.
    }
    \label{fig:fine_grained_image_weight}
\end{figure}

\begin{figure}
    \centering
    \begin{subfigure}[c]{0.48\linewidth}
        \centering
        \fbox{\includegraphics[width=\linewidth]{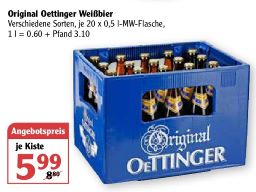}}
        \caption{Product by brand \textit{"Öttinger"}.}
    \end{subfigure}\hfill
    \begin{subfigure}[c]{0.48\linewidth}
        \centering
        \fbox{\includegraphics[width=\linewidth]{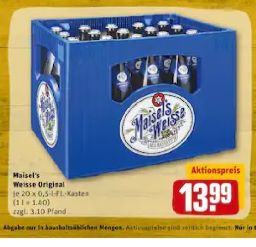}}
        \caption{Product by brand \textit{"Maisel's"}.}
    \end{subfigure}
    \caption{Illustrations of images that demonstrate a fine-grained difference due to the products are from different brands.
    The evaluation of ResNet50 \cite{he2016deep} reveals the misclassification of these images.
    }
    \label{fig:fine_grained_image_brand}
\end{figure}

\begin{figure}
    \centering
    \begin{subfigure}[c]{0.48\linewidth}
        \centering
        \fbox{\includegraphics[width=\linewidth]{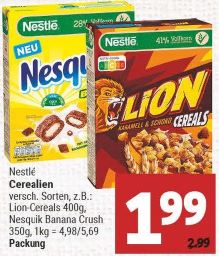}}
        \caption{The product is cereal.}
    \end{subfigure}\hfill
    \begin{subfigure}[c]{0.48\linewidth}
        \centering
        \fbox{\includegraphics[width=\linewidth]{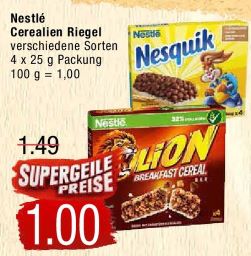}}
        \caption{The product is cereal bars.}
    \end{subfigure}
    \caption{Illustrations of images that demonstrate the fine-grained differences between the products. 
    The differences are in the products themselves, even though they are from the same brand.}
    \label{fig:fine_grained_image_products}
\end{figure}

The fine-grained characterization of the dataset is presented in a variety of forms.
Firstly, the assessment may depend on the image.
Advertisements of products with different weights illustrate a high visual similarity.
\Cref{fig:fine_grained_image_weight} shows such a similarity although the images belong to different classes due to the product weights.
Moreover, product advertisements for the same product but from different brands indicate visual similarity, too.
\Cref{fig:fine_grained_image_brand} illustrates this fine-grained emergence.
In addition, there are product advertisements that show different products of the same brand.
These promotions also demonstrate a fine-grained distinction.
\Cref{fig:fine_grained_image_products} illustrates advertisements of the different products, cereals and cereal bars.
The products derive from the same brand. 
Hence, the product packaging exhibit a strong visual similarity.\\
Even the product and promotion data manifest resemblance in the dataset.
\Cref{tab:fine_grained_text_weight} presents two items of the dataset that are only distinguished by the value of \textit{GTINs}, which is not visible in the corresponding image, and by the divergent value of \textit{product weight}.

\begin{table}
    \centering
    \begin{tabular}{lll}
        \hline
        target             &item 1             &item 2\\
        \hline
        brand               &Heinz              &Heinz\\
        product category    &Tomato Ketchup     &Tomato Ketchup\\
        GTINs               &08715700017006     &00000087157215\\
        product weight      &500.0 Milliliter   &400.0 Milliliter \\
        different sorts     &yes                &yes\\
        price               &1.99               &1.99\\
        regular price       &2.49               &2.49\\
        relative discount   &20                 &20\\
        absolute discount   &NaN                &NaN\\
        \hline
    \end{tabular}
    \caption{Illustration of the product and promotion data for two dataset items differing only in \textit{GTINs} and \textit{product weight}.}
    \label{tab:fine_grained_text_weight}
\end{table}
\section{Visual RAG Pipeline}
\label{sec:pipeline}
\noindent \Cref{fig:visual_abstract} illustrates the proposed Visual RAG pipeline, consisting of five main steps:
The first step is defined as \textit{Preprocessing} described in \Cref{subsec:preprocessing}.
In this step, the query image will be preprocessed to obtain the data that are stored in the \textit{Vector Store}.
The structure of the \textit{Vector Store} and its stored data is elucidated in \Cref{subsec:vector_store}.
The subsequent step is called as \textit{Retrieval, Classification, Relational Query}, which is outlined in \Cref{subsec:retrieval_classification}.
In this step, a retrieval is performed, followed by a classification being conducted using the retrieved data.
Based on the classification the step \textit{Relational Query} is executed.
In this step a context of few-shot samples is created.
The following step includes the \textit{Prompt Generation} presented in \Cref{subsec:prompt_generation}.
The last step of the pipeline is called \textit{Completion}, which executes a request to a VLM considering the generated prompt. A detailed description is defined in \Cref{subsec:completion}.
\begin{figure*}
    \centering
    \includegraphics[width=0.9\linewidth]{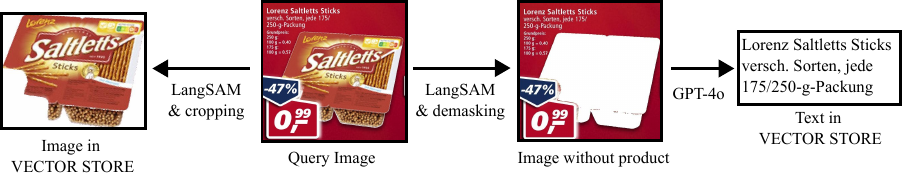}
    \caption{Illustration of the \textit{Preprocessing} step of the Visual RAG pipeline.
    The results of the \textit{Preprocessing} steps are stored in the \textit{Vector Store}.
    The images stored in \textit{Vector Store} are produced by the tool LangSAM \cite{medeiros2024langsegment} and image cropping.
    The result of using the tool LangSAM \cite{medeiros2024langsegment} and demasking is the input for the VLM GPT-4o \cite{OpenAI2024GPT4o} getting the product description.
    }
    \label{fig:pipeline_preprocessing}
\end{figure*}

\subsection{Preprocessing}
\label{subsec:preprocessing}
The input to the pipeline is an image as depicted in \Cref{subfig:example_image_dataset}.
The image is processed in two different ways depending on the data type stored in the \textit{Vector Store}.
Initially, the tool Language Segment Anything (LangSAM) \cite{medeiros2024langsegment} is applied to the query image. The result is a segmentation mask that is described by a given prompt. We use the prompt: \textit{"product."}.
For the image data that is stored in the \textit{Vector Store}, the 
segmentation mask is cropped from the query image.
The left image of \Cref{fig:pipeline_preprocessing} depicts the product image of the query image and is saved in the \textit{Vector Store}.
The product description printed on the advertisement image is also stored in the \textit{Vector Store}, but as text data. To obtain this text, another prepossessing step is necessary.
This time, the segmentation mask is eliminated from the query image, showing in \Cref{fig:pipeline_preprocessing} through the image \textit{"Image without product"}.
The VLM \textit{GPT-4o} \cite{OpenAI2024GPT4o} (version: 2024-08-06) is used for the extraction of the textual product description.
The system message is defined by: \textit{"You are an AI assistant that extract text from an image"}. The user prompt includes the preprocessed image without the product as well as the task description: \textit{"First, extract the text. Second, remove all price information. If available, remove all special / detailed description of the product"}.

\subsection{Vector Store}
\label{subsec:vector_store}
We use the Chroma vector database \cite{Chroma2024} as \textit{Vector Store} in our pipeline.
We add images and texts into \textit{Vector Store}.
The left image and the right text of \Cref{fig:pipeline_preprocessing} show exemplary data that are stored.
5,390 product images and 3,670 product description texts are stored in the \textit{Vector Store}.
The embedding vectors in the \textit{Vector Store} are created by the \textit{OpenCLIPEmbeddings} \cite{ilharco_gabriel_2021_5143773}.
For the text tokenizer, the model \textit{xlm-roberta-large-ViT-H-14} is used.
The similarity search in the \textit{Vector Store} is performed via cosine distance.
Each embedding vector stores additionally the information of the class label as meta data.

\subsection{Retrieval, Classification, Relational Query}
\label{subsec:retrieval_classification}
For the step \textit{Retrieval}, the five most similar embedding vectors to the query embedding are returned.
Each embedding vector has stored the class label.
Therefore, the following \textit{Classification} step determines the most frequently occurring class label.
In cases of ties, the class label of the nearest image embedding vector is returned.
Based on the additional label information, the embedding vectors can be filtered according to the classified label.
In the subsequent step \textit{Relational Query}, the corresponding advertisement image as well as the product and promotion data of the filtered embeddings vectors are delivered by an external database.
This information, images and texts, serves as few-shot samples and forms the contextual knowledge that is a part of the \textit{Prompt Generation}.

\subsection{Prompt Generation}
\label{subsec:prompt_generation}

The prompt given to the VLM is divided into three parts: task, query image, and context.
The task is formulated as: \textit{"Extract all features"}.
The features, also known as targets, refer to the product and promotion data.
As a further part of the prompt, a contextual knowledge is provided.
The context is understood as data consisting multiple few-shot samples, containing product images along with their corresponding product and promotion data.
The maximum number of samples is set to three, and a minimum of one sample is to be provided.
The number of samples depends on the VLM used.
Should the maximum input token length be exceeded by the samples, then the number of samples is reduced accordingly.
The number of data examples included in the context must be limited depending on the VLM due to the maximum input token length of the model.
The prompt is structured, beginning with the task description, followed by the contextual knowledge containing few-shot samples, and concluding with the query image.

\subsection{Completion}
\label{subsec:completion}

The \textit{Completion} step comprises the VLM request.
The input for the VLM is the prompt described in the previous section.
We investigate the VLMs: \textit{GPT-4o} \cite{OpenAI2024GPT4o} of the version 2024-08-06 (\textit{GPT-4o\_2024-08-06}), \textit{GPT-4o-mini} \cite{OpenAI_GPT4omini} of the version 2024-07-18 (\textit{GPT-4o-mini\_2024-07-18}), and \textit{Gemini-2.0-flash} \cite{googleGeminiModels}.
The models \textit{GPT-4o} and \textit{GPT-4o-mini} have a maximum input token length of 128,000 \cite{OpenAI2024GPT4o}.
These models offers the usage of structured output \cite{OpenAI2024} that is used.
The object schema is defined by using Pydantic \cite{pydanticWelcomePydantic}.
Each target is defined as an attribute of the Pydantic model with appropriate data types.
\Cref{appendix_sec:structured_output} in the Appendix shows the structured output of the VLM \textit{GPT-4o-mini\_2024-07-18} for the query image shown in \Cref{fig:example_data_of_dataset}.
\section{Evaluation}
\label{sec:evaluation}

In \Cref{subsec:eval_fine_grained_cls}, the various baseline evaluation procedures and the presented Visual RAG pipeline are investigated in relation to the FGC task.
Further examinations are described in 
\Cref{subsec:auxiliary_analysis}.

\subsection{Evaluation of Fine-Grained Classification}
\label{subsec:eval_fine_grained_cls}

For a baseline evaluation of the FGC task, we investigate image only, explained in \Cref{subsubsec:image_classification}, text only, described in \Cref{subsubsec:text_classification}, and combined image-text models, so-called multi-modal models, depicted in \Cref{subsubsec:multimodal_model}.
Moreover, the evaluation of the Visual RAG pipeline in relation to the FGC task is illustrated in \Cref{subsub:classification_visual_RAG_pipeline}.\\
\\
The examinations in \Cref{subsubsec:image_classification,subsubsec:text_classification} were conducted using a single NVIDIA Tesla T4 GPU.
The described investigations in \Cref{subsubsec:multimodal_model} and all evaluations of the Visual RAG pipeline are executed on a single CPU.\\
\\
The evaluation of the FGC task is based on the product target \textit{GTINs}.
A class of the dataset consists, among other things, of product advertisement images.
An image of a class can, \eg, advertise a single product, whilst other images of the same class promote multiple varieties of the same product.
Therefore, several GTINs are assigned to these images.
A class of the dataset is defined by the union of the target \textit{GTINs} values.
This forms the basis of the evaluation of the target \textit{GTINs} in \Cref{tab:evaluation_baseline_results}.


\subsubsection{Image Classification}
\label{subsubsec:image_classification}
For a baseline image-only classification, the model ResNet50 \cite{he2016deep} is used.
The configuration settings for the training are: epochs = 50, learning rate = 0.001, batch size = 32, optimizer = SGD, momentum of optimizer = 0.9, and image resize size = 256.
The duration of the training is almost two hours.
The evaluation of the trained model on the test dataset yields to an accuracy score of \textbf{84.4\%}.
Examples for misclassified images are shown in \Cref{fig:fine_grained_image_weight,fig:fine_grained_image_brand}.

\subsubsection{Text Classification}
\label{subsubsec:text_classification}
For a baseline text-only classification, the description text on the images is used.
We follow the text extraction method described in \cite{lamm2024visuallanguagemodelsreplace}, using
the OCR tool PyTesseract \cite{Hoffstaetter2014pytesseract}.
An image with the corresponding OCR-extracted product description text is included in \Cref{appendix_sec:ocr_extracted_text} in the Appendix.
The text classification is based on the model BERT \cite{kenton2019bert}.
The tokenizer and the classification model are provided by the Hugging Face transformers library \cite{huggingface}.
We use the model versions: \textit{bert-base-uncased} and \textit{BertForSequenceClassification}.
The configuration settings for the tokenizer and the classification model training are: tokenizer maximum length = 128, tokenizer padding = max length, epochs = 30, learning rate = 2e-5, batch size  32, and optimizer = AdamW.
The training duration is more than half an hour.
The evaluation of the trained model on the test dataset yields to an accuracy score of \textbf{74.2\%}.
Examples for false classified text shows the following:
OCR-extracted text of test data like \textit{"SÜDZUCKER  Puder Zucker Mühle* je 250-g-Dose  (100 Q = 0.40)"} or \textit{"tsmengen Un SÜDZUCKER Puder Zucker Mühle* je 250-g-Dose (100 g = 0.40)"} are misclassified into the class containing training data like \textit{"Diamant Gelierzucker 1:1 1-kg-Packung", "Diamant Gelierzucker In  E C TTT"}, and \textit{"Diamanb Gellerzucker  11 1-kg-Packung"}.
The classification by the OCR-extracted text of images, displayed in \Cref{subfig:fine_grained_image_weight_500}, are misclassified into the class, illustrated in \Cref{subfig:fine_grained_image_weight_400}.

\subsubsection{Multi-modal Classification with CLIP}
\label{subsubsec:multimodal_model}
The CLIP \cite{radford2021learning} model is based on the concept by learning from text-image pairs.
In particular, CLIP allows zero-shot classifications.
On the basis of these skills, we investigate the CLIP model as zero-shot baseline.
Specially, the model version \textit{ViT-B/32} is used.
In order to use zero-shot learning, it is necessary to provide a textual description of the classes.
In our case, the text is build by joining the prediction targets \textit{brand} and \textit{product category}.
The emphasis was placed on these two targets, as the \textit{brand} is typically visible in the image and the \textit{product category} constitutes an additional specification of the product.
The \textit{product category} can be utilized to differentiate between different products of the same \textit{brand}.
Moreover, the CLIP model \cite{radford2021learning} has a context length limitation.
As example for the image in \Cref{subfig:example_image_dataset} the class description is: \textit{"Lorenz - Saltletts Sticks"}.
The accuracy score of the CLIP model is about \textbf{45.9\%}.
This score is below the image and text classification.
However, there is neither training process nor costs associated with the usage of zero-shot learning on the CLIP model.

\subsubsection{Classification using our Visual RAG pipeline}
\label{subsub:classification_visual_RAG_pipeline}

The Visual RAG pipeline returns a prediction per each target.
This also applies to the target \textit{GTINs} according to which the FGC task is defined.
The classification using the Visual RAG pipeline is determined if the predicted \textit{GTINs} value is included in the GTINs union of a class.
Hence, a comparison of the baseline experiments can be made.
The examination of the Visual RAG pipeline using the VLM \textit{GPT-4o-mini\_2024-07-18} shows an accuracy score of \textbf{86.8\%}.\\
\Cref{tab:evaluation_baseline_results} shows the results of Image Classification, Text Classification, Multi-modal Classification, and Classification using the Visual RAG pipeline according to FGC.
The image-only classification with  ResNet50\cite{he2016deep} performs with an accuracy score of 84.4\%.
Text Classification and Multi-modal Classification are unable to approximate to the aforementioned value.
The accuracy scores are 74.5\% and 45.9\%, respectively.
The Classification using the Visual RAG pipeline outperforms the other methods by achieving an accuracy score of 86.8\%.
In addition, the Visual RAG pipeline provides a more comprehensive output which can be useful for further tasks.
The fact that there is no requirement for a model to be trained is a significant advantage of the Visual RAG pipeline.
There is a constant change of products in retail, so frequent training of models for image and text classification is necessary.
\begin{table}[!h]
    \centering
    \begin{tabular}{llc}
        \hline
        Classification  &Model      &GTINs\\
        \hline
        Baseline Image           &ResNet50\cite{he2016deep}      &84.4\%\\
        
        Baseline Text            &BERT\cite{kenton2019bert}      &74.5\%\\
        Baseline Multi-modal      &CLIP\cite{radford2021learning} &45.9\%\\
        \hline
        Visual RAG pipeline
        &\makecell[tl]{GPT-4o-mini\_\\2024-07-18\cite{OpenAI_GPT4omini}}
        &\textbf{86.8\%}\\
        \hline
        \end{tabular}
    \caption{Illustration of the accuracy scores of the FGC task represented by the GTINs.
    The evaluation of target \textit{GTINs} is based on 
    the set defined by the class, which is understood as the union of the GT values of this target.
    }
    \label{tab:evaluation_baseline_results}
\end{table}

\subsection{Auxiliary Analysis}
\label{subsec:auxiliary_analysis}
Further analysis pertaining to the Visual RAG pipeline are contemplated.
Specifically, the impact of the VLM and the contextual knowledge are discussed in \Cref{subsubsec:eval_different_vlms,subsubsec:eval_context}, respectively.
In \Cref{subsubsec:alternative_eval_measure} an alternative evaluation measure of the target \textit{GTINs} is described.
Subsequently, the evaluation of context biases and false predictions as well as the value and cost analysis follow in \Cref{subsub:eval_false_predictions,subsubsec:value_cost_analysis}.\\
The Visual RAG pipeline has been developed to make open-ended predictions across multiple product property targets, like \textit{price}, \textit{discount}, or \textit{brand} as introduced in \Cref{subsec:dataset_fine_grained}.
The method of validating the accuracy score of a target in the Visual RAG pipeline, depends on the target itself.
For the targets \textit{brand} and \textit{product category} it is sufficient if the predicted value is a substring of the Ground Truth (GT) value. Examples are as follows: the predicted \textit{brand} \textit{"LOreal"} is assessed correctly due to the GT value \textit{"[LOreal, Men Expert]"} as well as the predicted \textit{product category} \textit{"Pastasauce"} and the GT value \textit{"[Nudelsauce, Pasta Sauce, Pastasauce, Pasta-Sauce]"}.
The validity of this validation method is substantiated by the observation that the same product is advertised using different spellings for \textit{brand} and/or \textit{product category}.
For the other targets, the prediction has to be equal.
The requirement for identical \textit{GTINs} makes it possible to identify the promoted product on the advertisement images in a more specific way.

\subsubsection{Ablation Study: Impact of the VL-Model}
\label{subsubsec:eval_different_vlms}

The impact of VLMs in the Visual RAG pipeline is evaluated by incorporating the accuracy of all predicted targets.
The VLMs \textit{GPT-4o-mini\_2024-07-18} \cite{OpenAI_GPT4omini}, \textit{GPT-4o\_2024-08-06} \cite{OpenAI2024GPT4o}, and \textit{Gemini-2.0-flash} \cite{googleGeminiModels} are investigated.
\Cref{tab:evaluation_different_vlms} shows the accuracy score per target for the examined VLMs.
The variance in accuracy of the target \textit{GTINs} is significant.
The evaluation of this target is based on an equal numerical value of the prediction and GT.
The models \textit{Gemini-2.0-flash} and \textit{GPT-4o\_2024-08-06} reach 75.4\% and 76.5\% accuracy, respectively.
The model \textit{GPT-4o-mini\_2024-07-18} increases the accuracy score to 81.2\%.

\subsubsection{Ablation Study: Impact of contextual knowledge}
\label{subsubsec:eval_context}

The false predictions of the target \textit{GTINs} is explained in more detail.
In total, 206 of 1,101 test images receive a false predicted GTINs.
The most of these images receive a NULL value for the target \textit{GTINs}.
Due to the fact that the predictions for all other targets are also NULL, it suggests that the VLM response is invalid in these cases.
Further investigation has been conducted to deepen the understanding of the invalid response of the VLM.
The structure of the context containing few-shot samples is based on data similar to the query image.
The quantity of samples is predetermined at this stage.
In order to obtain a valid response, the context is reduced to contain only a single sample instead of multiple samples.
This procedure is applied using the VLM \textit{GPT-4o-mini\_2024-07-18}.
Hence, this approach achieves the best accuracy score of \textbf{86.0\%} for the target \textit{GTINs}, shown in the last column of \Cref{tab:evaluation_different_vlms}.
For this evaluation, the exact match of the GT value and prediction was applied instead of the metric used in \Cref{subsec:eval_fine_grained_cls}.


\begin{table*}[!htbp]
    \centering
    \begin{tabular}{llccc||c}
        \hline
        data type   &target
                    &\makecell{GPT-4o-mini\_\\2024-07-18}
                    &\makecell{GPT-4o\_\\2024-08-06}
                    &Gemini-2.0-flash
                    &\makecell{GPT-4o-mini\_\\2024-07-18$^+$}
                    \\
        \hline
        product     
        &brand              &90.8\%        &86.2\%       &90.7\%       &\textbf{96.8\%}\\
        &product category   &88.1\%        &82.6\%       &\textbf{93.0}\%       &\textbf{93.0\%}\\
        &product weight      &80.0\%        &77.7\%       &\textbf{85.2\%}       &84.7\%\\
        &GTINs        &81.2\% &76.5\% &75.4\% &\textbf{86.0\%}\\
        &different sorts     &\textbf{49.4\%}        &48.9\%       &49.0\%       &\textbf{49.4\%}\\
        promotion   
        &price               &87.9\%        &87.8\%       &91.5\%       &\textbf{91.8\%}\\
        &regular price       &36.5\%        &\textbf{49.2\%}       &47.6\%       &32.0\%\\
        &relative discount   &32.5\%        &\textbf{38.6\%}       &31.8\%       &30.4\%\\
        &absolute discount   &89.4\%        &\textbf{94.3\%}       &81.0\%       &88.2\%\\
        \hline  
    \end{tabular}
    \caption{Illustration of the accuracy scores per target for each examined VLM.
    The test dataset comprises of 1,101 images.
    The evaluation of target \textit{GTINs} is based on an equal numerical value of the prediction and GT.
    $^+$: If the VLM response exclusively contains NULL values then further requests with reduced context is executed.
    }
    \label{tab:evaluation_different_vlms}
\end{table*}


\subsubsection{Alternative evaluation measure}
\label{subsubsec:alternative_eval_measure}

So far, our evaluation for the target \textit{GTINs} requires predictions to be equal to the GT values, even though the GT value can comprise a list of GTINs.
A product advertisement often promotes multiple products using a single representative product image, \eg, different flavors of the same base product.
This results in numerous GTINs being assigned to the image.
A single correctly predicted GTIN should be recognized as a reference point.
For this reason, another evaluation method was applied.
The evaluation method is as follows: the prediction is considered as valid if at least a single predicted GTIN is correct.
According to this error analysis, the Visual RAG pipeline with the VLM \textit{GPT-4o-mini\_2024-07-18} using the method of reducing contextual knowledge elevates the accuracy score to \textbf{92.3\%}.

\begin{table*}[!htbp]
    \centering
    \begin{tabular}{lccc}
        \hline
        parameter   &GPT-4o-mini\_2024-07-18
                    &GPT-4o\_2024-08-06
                    &Gemini-2.0-flash\\
        \hline
        avg. input tokens           &92,888     &88,015     &104,119*\\
        avg. output tokens          &90         &87         &316*\\
        avg. total tokens           &92,978     &88,102     &104,436\\
        avg. elapsed time/req.[s]   &7.9        &10.2       &4.8\\
        approx. total costs         &\$15.28    &\$241.91   &\$10.98\\
        \hline
    \end{tabular}
    \caption{Illustration of various parameters for each examined VLM. *: Data of prompt and candidates tokens, respectively.
    }
    \label{tab:evaluation_different_vlms_parameters}
\end{table*}

\subsubsection{Context Biases and False Predictions}
\label{subsub:eval_false_predictions}
\begin{figure}
    \centering
    \begin{minipage}{0.4\linewidth}
        \fbox{\includegraphics[width=\linewidth]{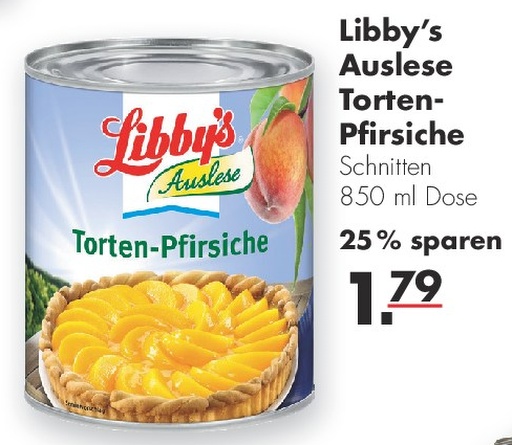}}
        \label{subfig:evaluation_image_GTIN_Pfirsich}
    \end{minipage}
    \hfill
    \begin{minipage}{0.48\linewidth}
        \small 
        \begin{tabular}{ll}
            \hline
                            &GTINs\\
            \hline
            GT              &07613034229083\\ 
            prediction      &07613034228673,\\
                            &07613034228826,\\
                            &07613034229083\\
            \hline
        \end{tabular}
        \label{subfig:evaluation_text_GTIN_Pfirsich}
    \end{minipage}
    \caption{Illustration of an image plus GT and prediction values for the target \textit{GTINs}.
    The prediction comprises a list of GTINs due to the fact that the data in the context exclusively promote for different sorts of the product.}
    \label{fig:evaluation_GTIN_Pfirsich}
\end{figure}

Introducing few-shot context into the prediction pipeline also has some drawbacks, as biased context can lead to false predictions. In the following, we summarize context biases we could observe in our evaluation. Please refer to \Cref{appendix_sec:false_prediction} in the Appendix for visualizations of the described cases.
\\\noindent\textbf{Image-Text Bias.} In some cases, the models ignore the correct contextual information regarding the GTIN and extract arbitrary numbers which are visible in the query image instead. \Cref{fig:evaluation_GTIN_Gewuerzgurke} in the Appendix shows an example for this behavior. A possible mitigation for this bias is to specifically instruct the VLM not to use printed numbers in the query image to predict the GTIN.  
\\\noindent\textbf{Image-Context Bias.} \Cref{fig:evaluation_GTIN_Pfirsich} shows an advertisement image that promotes only a single product. The hint that the price is valid for different sorts is not printed in the image.
Hence, the GT value for the target \textit{GTINs} only consists of a single number.
However, the context of few-shot samples provided for the VLM contained images promoting different sorts of the product.
Therefore, the product data of the context comprise a list of GTINs per each context image.
\\
Basically, the quality of the GT values for the target \textit{GTINs} is considered critically.
There are images in the dataset that promote a single product without the additional note of validity for different sorts.
However, the product target \textit{GTINs} contains a list of numbers.
Also, the opposite appears: the advertisement image promotes a product in different sorts but only a single GTIN is stored in the data.
\\\noindent\textbf{Database Bias.}
Due to marketing impacts, some advertisements have displayed more than one price in the image.
It is also the case that the printed pricing information is not the unit of the promoted product.
For example, an advertisement promotes for a carton of six bottles.
However, the price detail is per bottle and the GT value in the dataset is defined as price for the carton with six bottles.
\\\noindent\textbf{Missing Data Bias.}
If the advertisement image does not display the information of regular price, the prediction is taken from the context containing few-shot samples of similar data.
Further, the indication of the recommended retail price is false predicted as regular price.
The recommended retail price is basically not defined as the GT for the target \textit{regular price}.
Other false predictions for the named target include the declaration about the reference weight price.
A false prediction of this kind happens especially if there is no specification about regular price in the image.

\subsubsection{Value and Cost Analysis}
\label{subsubsec:value_cost_analysis}
The cost analysis depends on the token numbers.
The number of FLOPs is not specifiable as only commercial models are used.
All steps of the Visual RAG pipeline are executed on a single CPU.
The average duration of the \textit{Completion} step of the VLM \textit{GPT-4o-mini\_2024-07-18} is about 7.9 seconds.
The average costs per a single \textit{Completion} step is about \$0.01.
Hence, the total costs for the test set amounts to \$15.28.
Further costs may arise due to the allocation of the VLM and the RAG database.
\Cref{tab:evaluation_different_vlms_parameters} provides an overview of various parameters for each VLM investigated. 
\section{Limitations and Outlook}
\label{sec:limitations_outlook}
\noindent This paper introduces a dataset consisting of images as well as product and promotion data per image.
With the definition of the prediction target \textit{GTINs}, the dataset is investigated for the task of FGC.
Furthermore, we present a Visual RAG pipeline containing of five main steps.
The results of the \textit{Preprocessing} step serve as input for the \textit{Vector Store}.
The composition of the context that consists of few-shot samples (image, product and promotion data) is produced by the step \textit{Retrieval, Classification, Relational Query}.
The \textit{Prompt Generation} step joins the query image, the context, and the task description.
The result of the pipeline or the output of the \textit{Completion} delivers the product and promotion data for the query image.
The efficacy of the entire pipeline is contingent upon the quality of the segmentation mask, which is the result of the LangSAM tool \cite{medeiros2024langsegment}.
The segmentation of the product on the image can fail.
Hence, either only a part of the product or no mask is segmented.
Image examples are included in \Cref{appendix_sec:defect_prod_seg_exp} in the Appendix.
Moreover, the impact of the embedding model and the \textit{Vector Store} type can be investigated in more depth.
Further, reducing the context in term of the number of samples enables a valid response from the VLM used.
Hence, the reduction of the context in relation to the image resolution can be investigated.
Consequently, the printed text on images may then be no longer legible.
In addition, novel rapidly emerging VLMs and new findings in term of Prompt Engineering can be analyzed.
The analysis of different VLMs used in the Visual RAG pipeline shows that the model has significant influence of the result quality.
This also means that Visual RAG pipeline is limited due to the external VLMs.
The significant advantage of the presented Visual RAG pipeline is the minimal effort for new/unknown retail products.
In contrast to the training of an image/text classifier, only few-shot samples of such product has to be provided in the \textit{Vector Store}.
In \Cref{sec:evaluation}, the error analysis demonstrates the limitations.
Numerous false predictions for the target \textit{GTINs} are based to the insufficient data quality.
Therefore, data cleaning, especially focusing on the product data \textit{GTINs}, is essential.
Consequently, several more investigations on the dataset and the Visual RAG pipeline may endure.
\newpage
{
    \small
    \bibliographystyle{ieeenat_fullname}
    \bibliography{main}
}

\clearpage
\maketitlesupplementary
\noindent Additional analyses can be found in the supplementary \Cref{appendix_sec:ocr_extracted_text,appendix_sec:structured_output,appendix_sec:false_prediction,appendix_sec:defect_prod_seg_exp}.
\Cref{appendix_sec:ocr_extracted_text} shows an example of OCR-extracted text used by the Fine-Grained Classification (FGC) task solved by Text Classification, see \Cref{subsubsec:text_classification}.
The structured output of a VLM response is depicted in \Cref{appendix_sec:structured_output}.
Further illustrations of erroneous predictions yielded by the Visual RAG pipeline are presented in \Cref{appendix_sec:false_prediction}.
\Cref{appendix_sec:defect_prod_seg_exp} includes examples of defective product segmentation masks. The segmentation masks are necessary for the initial step \textit{Preprocessing} in the Visual RAG pipeline.

\section{OCR-extracted Text}
\label{appendix_sec:ocr_extracted_text}
\begin{figure}[!ht]
    \centering
    \begin{minipage}{0.48\linewidth}
        \centering
        \fbox{\includegraphics[width=\linewidth]{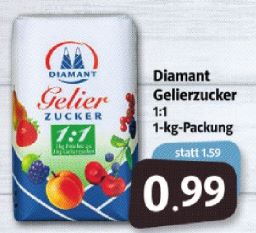}}
        \label{subfig:text_classification_image}
    \end{minipage}
    \hfill
    \begin{minipage}{0.5\linewidth}
        \small 
        Diamant  Gelierzucker 1:1 1-kg-Packung
\label{subfig:text_classification_product_description_text}
    \end{minipage}
    \caption{Image and its OCR-extracted product description text.}
    \label{fig:text_classification_example}
\end{figure}

\Cref{fig:text_classification_example} shows an image with the corresponding OCR-extracted product description text by using PyTesseract \cite{Hoffstaetter2014pytesseract}.

\section{Structured Output of VLM Response}
\label{appendix_sec:structured_output}

The following is the structured output of the VLM \textit{GPT-4o-mini\_2024-07-18} for the query image shown in \Cref{fig:example_data_of_dataset}:
\begin{verbatim}
brand='Lorenz'
price=0.99
regular_price=1.87
relative_discount=47
absolute_discount=None
product_category=['Saltletts Sticks']
GTINs=['04018077683015',
       '04018077686719']
weight_number=250.0
weight_unit=Gramm
different_sorts=yes
\end{verbatim}

\section{Examples of False Predictions}
\label{appendix_sec:false_prediction}

\begin{figure}[!ht]
    \centering
    \begin{minipage}{0.48\linewidth}
        \centering
        \fbox{\includegraphics[width=\linewidth]{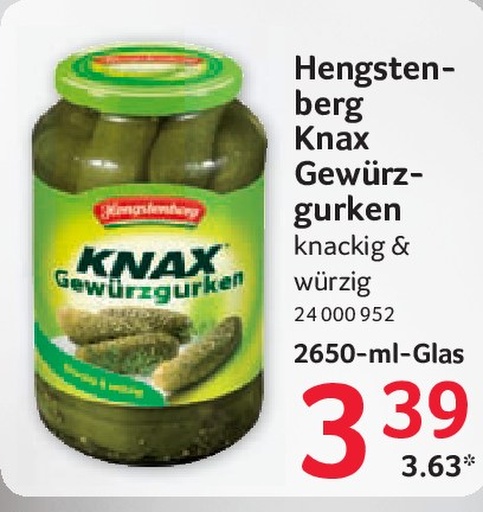}}
        \label{subfig:evaluation_image_GTIN_Gewuerzgurke}
    \end{minipage}
    \hfill
    \begin{minipage}{0.48\linewidth}
        \small 
        \begin{tabular}{ll}
            \hline
                            &GTINs\\
            \hline
            GT             &04008100140301\\ 
            prediction     &24000952\\
            \hline
        \end{tabular}
        \label{subfig:evaluation_text_GTIN_Gewuerzgurke}
    \end{minipage}
    \caption{Illustration of an image plus GT and prediction values for the target \textit{GTINs}.
    The false prediction is based on the extraction of text from the image.}
    \label{fig:evaluation_GTIN_Gewuerzgurke}
\end{figure}

\begin{figure}[!ht]
    \centering
    \begin{subfigure}[c]{\linewidth}
        \centering
        \fbox{\includegraphics[width=\linewidth]{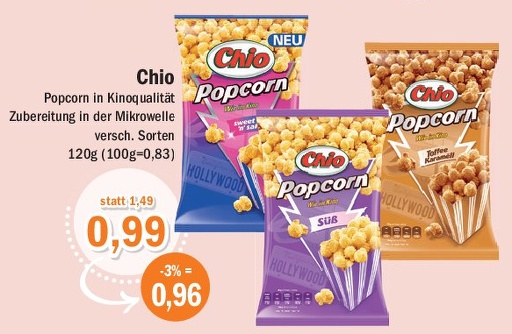}}
        \caption{Multiple price information.}
        \label{subfig:evaluation_multiple_price_info}
    \end{subfigure}\hfill
    \begin{subfigure}[c]{0.7\linewidth}
        \centering
        \fbox{\includegraphics[width=\linewidth]{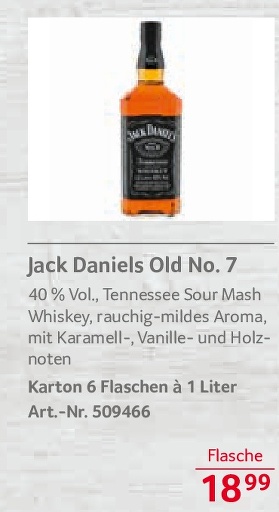}}
        \caption{Price information per bottle.}
        \label{subfig:evaluation_price_bottle_carton}
    \end{subfigure}
    \caption{Illustration of images for which the price predictions are false.
    The left image contains two information for the price: 0.99€ and a reduced price of 0.96€.
    The printed price information in the right image is the price per bottle.
    But the advertisement promotes a carton containing six bottles.
    }
    \label{fig:evaluation_prices}
\end{figure}

\begin{figure}[ht]
    \centering
    \begin{subfigure}[c]{0.7\linewidth}
        \centering
        \fbox{\includegraphics[width=\linewidth]{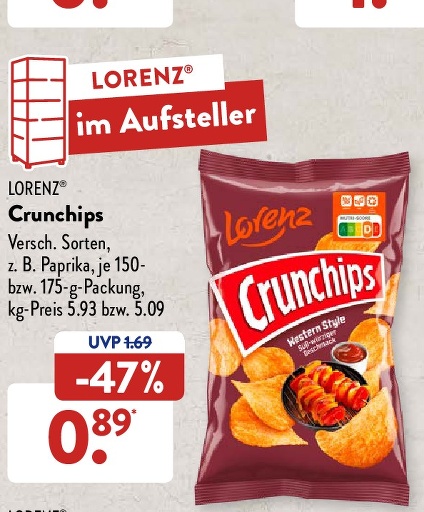}}
        \caption{Recommended retail price.}
        \label{subfig:evaluation_recommended_retail_price}
    \end{subfigure}\hfill
    \begin{subfigure}[c]{\linewidth}
        \centering
        \fbox{\includegraphics[width=\linewidth]{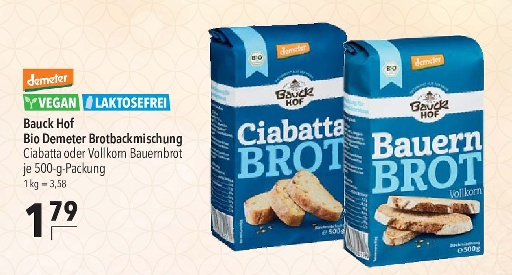}}
        \caption{Reference weight price.}
        \label{subfig:evaluation_reference_weight_price}
    \end{subfigure}
    \caption{Illustration of images for which the regular price predictions are false.
    \Cref{subfig:evaluation_recommended_retail_price} shows the print of the recommended retail price.
    The reference weight price of 1kg is 3.58€ for the advertisement image in \Cref{subfig:evaluation_reference_weight_price}.
    }
    \label{fig:evaluation_regular_prices}
\end{figure}

\Cref{fig:evaluation_GTIN_Gewuerzgurke} shows an image with the GT value and prediction for the target \textit{GTINs}.
The false prediction is based on the extraction of text from the image.
\Cref{subfig:evaluation_multiple_price_info,subfig:evaluation_price_bottle_carton} show images for which false predictions for target \textit{price} are resulted from the Visual RAG pipeline.
Regarding the false predictions for the target \textit{regular price}, 
\Cref{subfig:evaluation_recommended_retail_price,subfig:evaluation_reference_weight_price,} display in each case an image.

\section{Defect Product Segmentation Examples}
\label{appendix_sec:defect_prod_seg_exp}
\begin{figure}[ht]
    \centering
    \fbox{\includegraphics[width=0.8\linewidth]{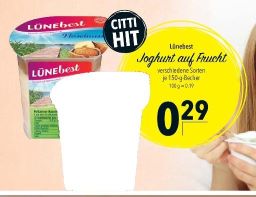}}
    \caption{Missing second printed product image for the segmentation mask.}
    \label{fig:langsam_missing_product}
\end{figure}

\begin{figure}[h]
    \centering
    \fbox{\includegraphics[width=0.8\linewidth]{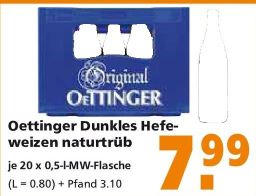}}
    \caption{Segmentation mask includes the bottles without the beer crate.}
    \label{fig:langsam_beer_crate}
\end{figure}

\begin{figure}[!t]
    \centering
    \fbox{\includegraphics[width=0.8\linewidth]{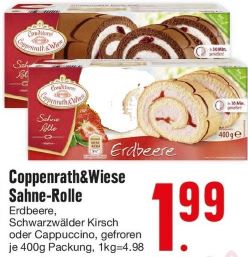}}
    \caption{Only a part of the product image is segmented.}
    \label{fig:langsam_part_of_packaing}
\end{figure}

\Cref{fig:langsam_missing_product} shows an image without the second segmented product.
The result shows that only one product image is segmented.
The second printed product image is not found.
\Cref{fig:langsam_beer_crate,fig:langsam_part_of_packaing} show images in which only a part of the actual product is segmented.
In \Cref{fig:langsam_beer_crate} only the bottles without the beer crate is segmented.
In \Cref{fig:langsam_part_of_packaing}, only a part of the product image is the segmentation mask.

\end{document}